\documentclass[11pt,a4paper]{article}
\usepackage[hyperref]{emnlp2020}
\usepackage{times}
\usepackage{latexsym}

\usepackage{microtype}

\aclfinalcopy 
\def\aclpaperid{119} 

\usepackage{graphicx}
\usepackage{todonotes}
\usepackage{amsmath}
\usepackage{mathtools}
\usepackage{url}
\usepackage{caption, subcaption}
\usepackage{lipsum}

\usepackage{color}
\usepackage{tcolorbox}
\usepackage{CJK}
\usepackage{adjustbox}
\usepackage[nodisplayskipstretch]{setspace}

\newcommand\Tstrut{\rule{0pt}{2.6ex}}         
\newcommand\Bstrut{\rule[-0.9ex]{0pt}{0pt}}   

\newcommand\blfootnote[1]{%
  \begingroup
  \renewcommand\thefootnote{}\footnote{#1}%
  \addtocounter{footnote}{-1}%
  \endgroup
}

\title{MAST: Multimodal Abstractive Summarization with Trimodal Hierarchical Attention}

\author{Aman Khullar$^*$ \\
  IIIT Hyderabad \\
  Hyderabad, India \\
  {\tt aman.khullar@iiit.ac.in} \\\And
  Udit Arora$^*$ \\
  New York University \\
  New York, NY, USA \\
  {\tt uditarora@nyu.edu} \\}

\date{}

\begin{document}
\maketitle

\begin{abstract}
  \blfootnote{* indicates equal contribution}
  \blfootnote{Aman Khullar is presently at Gram Vaani}
  \blfootnote{This paper will appear at the first EMNLP Workshop}\blfootnote{on NLP Beyond Text, 2020}
  This paper presents MAST, a new model for Multimodal Abstractive Text Summarization that utilizes information from all three modalities -- text, audio and video -- in a multimodal video. Prior work on multimodal abstractive text summarization only utilized information from the text and video modalities. We examine the usefulness and challenges of deriving information from the audio modality and present a sequence-to-sequence trimodal hierarchical attention-based model that overcomes these challenges by letting the model pay more attention to the text modality. MAST outperforms the current state of the art model (video-text) by 2.51 points in terms of Content F1 score and 1.00 points in terms of Rouge-L score on the How2 dataset for multimodal language understanding.
  
\end{abstract}

\section{Introduction}
In recent years, there has been a dramatic rise in information access through videos, facilitated by a proportional increase in the number of video-sharing platforms. This has led to an enormous amount of information accessible to help with our day-to-day activities. The accompanying transcripts or the automatic speech-to-text transcripts for these videos present the same information in the textual modality. However, all this information is often lengthy and sometimes incomprehensible because of verbosity. These limitations in user experience and information access are improved upon by the recent advancements in the field of multimodal text summarization.

Multimodal text summarization is the task of condensing this information from the interacting modalities into an output summary. This generated output summary may be unimodal or multimodal \cite{zhu2018msmo}. The textual summary may, in turn, be extractive or abstractive. The task of extractive multimodal text summarization involves selection and concatenation of the most important sentences in the input text without altering the sentences or their sequence in any way. \citet{li2017multi} made the selection of these important sentences using visual and acoustic cues from the corresponding visual and auditory modalities. On the other hand, the task of abstractive multimodal text summarization involves identification of the theme of the input data and the generation of words based on the deeper understanding of the material. This is a tougher problem to solve which has been alleviated with the advancements in the abstractive text summarization techniques -- \citet{rush2015neural}, \citet{see2017get} and \citet{liu2019text}. \citet{sanabria2018how2} introduced the How2 dataset for large-scale multimodal language understanding, and \citet{palaskar2019multimodal} were able to produce state of the art results for multimodal abstractive text summarization on the dataset. They utilized a sequence-to-sequence hierarchical attention based technique \cite{libovicky2017attention} for combining textual and image features to produce the textual summary from the multimodal input. Moreover, they used speech for generating the speech-to-text transcriptions using pre-trained speech recognizers, however it did not supplement the other modalities.
\begin{table}
    \centering
    \footnotesize
    \begin{tabular}{|p{0.95\linewidth}|}
        \hline
        \Tstrut
        \textbf{Original text:} let’s talk now about how \textcolor{red}{to} bait \textcolor{red}{a} tip up hook with \textcolor{red}{a} maggot. typically, \textcolor{red}{you}'re \textcolor{red}{going} \textcolor{red}{to} be using \textcolor{red}{this} for pan fish. not \textcolor{red}{a} real well known or common technique but \textcolor{red}{on} \textcolor{red}{a} given day \textcolor{red}{it} could be \textcolor{red}{the} difference between not catching fish \textcolor{red}{and} catching fish. all \textcolor{red}{you} do, \textcolor{red}{you} take \textcolor{red}{your} maggot, \textcolor{red}{you} can use meal worms, as well, which are much bigger, which are probably more well suited for \textcolor{red}{this} because \textcolor{red}{this} \textcolor{red}{is} \textcolor{red}{a} rather large hook. \textcolor{red}{you} would just, again, put \textcolor{red}{that} hook right through \textcolor{red}{the} maggot. with \textcolor{red}{a} big hook like \textcolor{red}{this}, \textcolor{red}{i} would probably put ten \textcolor{red}{of} these \textcolor{red}{on} \textcolor{red}{it}, just line \textcolor{red}{the} whole thing. \textcolor{red}{this} \textcolor{red}{is} \textcolor{red}{going} \textcolor{red}{to} be more \textcolor{red}{of} \textcolor{red}{a} technique for pan fish, such as, perch \textcolor{red}{and} sunfish, some \textcolor{red}{of} \textcolor{red}{your} smaller fish but if \textcolor{red}{you} had maggots, like \textcolor{red}{this} , or \textcolor{red}{a} meal worm, or two, \textcolor{red}{on} \textcolor{red}{a} hook like \textcolor{red}{this}, \textcolor{red}{this} would be \textcolor{red}{a} fantastic setup for trout, as well.\Bstrut\\
        \hline
        \Tstrut
        \textbf{Text only:} ice fishing is used \textcolor{red}{for} ice fishing. \textcolor{red}{learn} about ice fishing bait \textcolor{red}{with} \textcolor{red}{tips} \textcolor{red}{from} \textcolor{red}{an} experienced fisherman artist \textcolor{red}{in} \textcolor{red}{this} \textcolor{red}{free} fishing \textcolor{red}{video}.\Bstrut\\
        \hline
        \Tstrut
        \textbf{Video-Text:} \textcolor{red}{learn} about \textcolor{red}{the} ice fishing bait \textcolor{red}{in} \textcolor{red}{this} ice fishing lesson \textcolor{red}{from} \textcolor{red}{an} experienced fisherman.\Bstrut\\
        \hline
        \Tstrut
        \textbf{MAST:} maggots are good \textcolor{red}{for} catching perch. \textcolor{red}{learn} more about ice fishing bait \textcolor{red}{in} \textcolor{red}{this} ice fishing lesson \textcolor{red}{from} \textcolor{red}{an} experienced fisherman.\Bstrut\\
        \hline
    \end{tabular}
    \caption{Comparison of outputs by using different modality configurations for a test video example. Frequently occurring words are highlighted in \textcolor{red}{red}, which are easier for a simpler model to predict but do not contribute much in terms of useful content. The summary generated by the MAST model contains more content words as compared to the baselines.}
    \label{tab:contents}
    \vspace{-3.5mm}
\end{table}

Though the previous work in abstractive multimodal text summarization has been promising, it has not yet been able to capture the effects of combining the audio features. Our work improves upon this shortcoming by examining the benefits and challenges of introducing the audio modality as part of our solution. We hypothesize that the audio modality can impart additional useful information for the text summarization task by letting the model pay more attention to words that are spoken with a certain tone or level of emphasis. Through our experiments, we were able to prove that not all modalities contribute equally to the output. We found a higher contribution of text, followed by video and then by audio. This formed the motivation for our MAST model, which places higher importance on text input while generating the output summary. MAST is able to produce a more illustrative summary of the original text (see Table \ref{tab:contents}) and achieves state of the art results.

In summary, our primary contributions are:
\vspace{-2.5mm}
\begin{itemize}
    \setlength\itemsep{0.05pt}
    \item Introduction of audio modality for abstractive multimodal text summarization.
    \item Examining the challenges of utilizing audio information and understanding its contribution in the generated summary.
    \item Proposition of a novel state of the art model, MAST, for the task of multimodal abstractive text summarization.
\end{itemize}

\section{Methodology}
In this section we describe (1) the dataset used, (2) the modalities, and (3) our MAST model's architecture. The code for our model is available online\footnote{\url{https://github.com/amankhullar/mast}}.
\subsection{Dataset}
We use the 300h version of the How2 dataset \cite{sanabria2018how2} of open-domain videos. The dataset consists of about 300 hours of short instructional videos spanning different domains such as cooking, sports, indoor/outdoor activities, music, and more. A human-generated transcript accompanies each video, and a 2 to 3 sentence summary is available for every video, written to generate interest in a potential viewer. The 300h version is used instead of the 2000h version because the audio modality information is only available for the 300h subset.

The dataset is divided into the training, validation and test sets. The training set consists of 13,168 videos totaling 298.2 hours. The validation set consists of 150 videos totaling 3.2 hours, and the test set consists of 175 videos totaling 3.7 hours. A more detailed description of the dataset has been given by \citet{sanabria2018how2}. For our experiments, we took 12,798 videos for the training set, 520 videos for the validation set and 127 videos for the test set.

\subsection{Modalities}
We use the following three inputs corresponding to the three different modalities used:
\vspace{-2.5mm}
\begin{itemize}
    \setlength\itemsep{0.05pt}
    \item \textbf{Audio}: We use the concatenation of 40-dimensional Kaldi \cite{povey2011kaldi} filter bank features from 16kHz raw audio using a time window of 25ms with 10ms frame shift and the 3-dimensional pitch features extracted from the dataset to obtain the final sequence of 43-dimensional audio features.
    \item \textbf{Text}: We use the transcripts corresponding to each video. All texts are normalized and lower-cased.
    \item \textbf{Video}: We use a 2048-dimensional feature vector per group of 16 frames, which is extracted from the videos using a ResNeXt-101 3D CNN trained to recognize 400 different actions \cite{hara2018can}. This results in a sequence of feature vectors per video.
\end{itemize}

\subsection{Multimodal Abstractive Summarization with Trimodal Hierarchical Attention}
\begin{figure}
    \centering
    \includegraphics[scale=0.28]{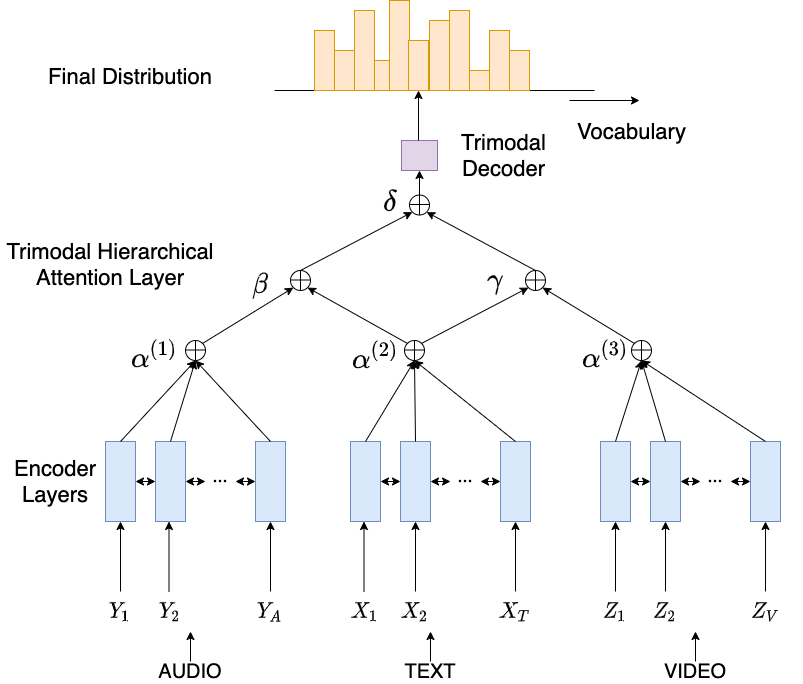}
    \caption{Multimodal Abstractive Summarization with Trimodal Hierarchical Attention (MAST) architecture: MAST is a sequence to sequence model that uses information from all three modalities -- audio, text and video. The modality information is encoded using Modality Encoders, followed by a Trimodal Hierarchical Attention Layer, which combines this information using a three-level hierarchical attention approach. It attends to two pairs of modalities ($\delta$) (Audio-Text and Video-Text) followed by the modality in each pair ($\beta$ and $\gamma$), followed by the individual features within each modality ($\alpha$). The decoder utilizes this combination of modalities to generate the output over the vocabulary.}
    \label{fig:trimodal_arch}
    \vspace{-3.5mm}
\end{figure}
Figure \ref{fig:trimodal_arch} shows the architecture of our Multimodal Abstractive Summarization with Trimodal Hierarchical Attention (MAST) model. The model consists of three components - Modality Encoders, Trimodal Hierarchical Attention Layer and the Trimodal Decoder.

\subsubsection{Modality Encoders}
The text is embedded with an embedding layer and encoded using a bidirectional GRU encoder. The audio and video features are encoded using bidirectional LSTM encoders. This gives us the individual output encoding corresponding to all modalities at each encoder timestep. The tokens $t_i^{(k)}$ corresponding to modality $k$ are encoded using the corresponding modality encoders and produce a sequence of hidden states $h_i^{(k)}$ for each encoder time step ($i$).

\subsubsection{Trimodal Hierarchical Attention Layer}
We build upon the hierarchical attention approach proposed by \citet{libovicky2017attention} to combine the modalities. On each decoder timestep $i$, the attention distribution ($\alpha$) and the context vector for the $k$-th modality is first computed independently as in \citet{bahdanau2014neural}:
\begin{gather}
    e_{ij}^{(k)} = v_a^{(k)\text{T}} \text{tanh}(W_a^{(k)} s_i + U_a^{(k)} h_j^{(k)} + b_{att}^{(k)}) \\
    \alpha_{ij}^{(k)} = \text{softmax}(e_{ij}^{(k)}) \label{eq2}\\
    c_i^{(k)} = \sum_{\mathclap{\substack{j=1,\\k \in \{\text{audio, text, video}\}}}}^{N_k} \alpha_{ij}^{(k)} h_j^{(k)}
\end{gather}
Where $s_i$ is the decoder hidden state at $i$-th decoder timestep, $h_j^{(k)}$ is the encoder hidden state at $j$-th encoder timestep, $N_k$ is the number of encoder timesteps for the $k$-th modality and $e_{ij}^{(k)}$ is attention energy corresponding to them. $W_a$ and $U_a$ are trainable projection matrices, $v_a$ is a weight vector and $b_{att}$ is the bias term.

We now look at two different strategies of combining information from the modalities. The first is a simple extension of the hierarchical attention combination. The second is the strategy used in MAST, which combines modalities using three levels of hierarchical attention.

\textbf{1. TrimodalH2}:
To obtain our first baseline model (\emph{TrimodalH2}), with 2 level attention hierarchy, the context vectors for all three modalities are combined using a second layer of attention mechanism and its context vector is computed separately by using hierarchical attention combination as in \citet*{libovicky2017attention}:
\begin{gather}
    e_i^{(k)} = v_b^T \text{tanh}(W_b s_i + U_b^{(k)}c_i^{(k)}) \\
    \eta_i^{(k)} = \text{softmax}(e_i^{(k)}) \\
    c_i = \sum_{\mathclap{k \in \{\text{audio, text, video}\}}} \eta_i^{(k)} U_c^{(k)} c_i^{(k)}
\end{gather}
where $\eta^{(k)}$ is the hierarchical attention distribution over the modalities, $c_i^{(k)}$ is the context vector of the $k$-th modality encoder, $v_b$ and $W_b$ are shared parameters across modalities, and $U_b^{(k)}$ and $U_c^{(k)}$ are modality-specific projection matrices.

\textbf{2. MAST}:
To obtain our MAST model, the context vectors for audio-text and text-video are combined using a second layer of hierarchical attention mechanisms ($\beta$ and $\gamma$) and their context vectors are computed separately. These context-vectors are then combined using the third hierarchical attention mechanism ($\delta$).

\emph{1. Audio-Text}:
\begin{gather}
    e_i^{(k)} = v_d^T \text{tanh}(W_d s_i + U_d^{(k)}c_i^{(k)}) \\
    \beta_i^{(k)} = \text{softmax}(e_i^{(k)}) \label{eq8}\\
    d_i^{(1)} = \sum_{\mathclap{k \in \{\text{audio, text}\}}} \beta_i^{(k)} U_e^{(k)} c_i^{(k)}
\end{gather}
\hspace{3mm} \emph{2. Video-Text}:
\begin{gather}
    e_i^{(k)} = v_f^T \text{tanh}(W_f s_i + U_f^{(k)}c_i^{(k)}) \\
    \gamma_i^{(k)} = \text{softmax}(e_i^{(k)}) \label{eq11} \\
    d_i^{(2)} = \sum_{\mathclap{k \in \{\text{video, text}\}}} \gamma_i^{(k)} U_g^{(k)} c_i^{(k)}
\end{gather}
where $d_i^{(l)}$, $l \in \{$audio-text, video-text$\}$ is the context vector obtained for the corresponding pair-wise modality combination.

Finally, these audio-text and video-text context vectors are combined using the third and final attention layer ($\delta$). With this trimodal hierarchical attention architecture, we combine the textual modality twice with the other two modalities in a pair-wise manner, and this allows the model to pay more attention to the textual modality while incorporating the benefits of the other two modalities.
\begin{gather}
    e_i^{(l)} = v_h^T \text{tanh}(W_g s_i + U_h^{(l)}d_i^{(l)}) \\
    \delta_i^{(l)} = \text{softmax}(e_i^{(l)}) \label{eq14}\\
    c_i^f = \sum_{\mathclap{l \in \{\text{audio-text, video-text}\}}} \delta_i^{(l)} U_m^{(l)} d_i^{(l)}
\end{gather}
where $c_i^f$ is the final context vector at $i$-th decoder timestep.

\subsubsection{Trimodal Decoder}
We use a GRU-based conditional decoder \cite{firat2016git} to generate the final vocabulary distribution at each timestep. At each timestep, the decoder has the aggregate information from all the modalities. The trimodal decoder focuses on the modality combination, followed by the individual modality, then focuses on the particular information inside that modality. Finally, it uses this information along with information from previous timesteps, which is passed on to two linear layers to generate the next word from the vocabulary.

\section{Experiments}
We train Trimodal Hierarchical Attention (MAST) and TrimodalH2 models on the 300h version of the How2 dataset, using all three modalities. We also train Hierarchical Attention models considering Audio-Text and Video-Text modalities, as well as simple Seq2Seq models with attention for each modality individually as baselines. As observed by \citet{palaskar2019multimodal}, the Pointer Generator model \cite{see2017get} does not perform as well as Seq2Seq models on this dataset, hence we do not use that as a baseline in our experiments. We consider another transformer-based baseline for the text modality, BertSumAbs \cite{liu2019text}.

For all our experiments (except for the BerSumAbs baseline), we use the \emph{nmtpytorch} toolkit \cite{caglayan2017nmtpy}. The source and the target vocabulary consists of 49,329 words on which we train our word embeddings. We use the NLL loss and the Adam optimizer \cite{kingma2014adam} with learning rate 0.0004 and trained the models for 50 epochs. We generate our summaries using beam search with a beam size of 5, and then evaluate them using the ROUGE metric \cite{lin2004rouge} and the Content F1 metric \cite{palaskar2019multimodal}.

In our experiments, the text is embedded with an embedding layer of size 256 and then encoded using a bidirectional GRU encoder \cite{cho2014learning} with a hidden layer of size 128, which gives us a 256-dimensional output encoding corresponding to the text at each timestep. The audio and video frames are encoded using bidirectional LSTM encoders \cite{hochreiter1997long} with a hidden layer of size 128, which gives a 256-dimensional output encoding corresponding to the audio and video features at each timestep. Finally, the GRU-based conditional decoder uses a hidden layer of size 128 followed by two linear layers which transform the decoder output to generate the final output vocabulary distribution.

To improve generalization of our model, we use two dropout layers within the Text Encoder and one dropout layer on the output of the conditional decoder, all with a probability of 0.35. We also use implicit regularization by using early stopping mechanism on the validation loss with a patience of 40 epochs.

\subsection{Challenges of using audio modality}
The first challenge comes with obtaining a good representation of the audio modality that adds value beyond the text modality for the task of text summarization. As found by \citet*{Mohamed2014DeepNN}, DNN acoustic models prefer features that smoothly change both in time and frequency, like the log mel-frequency spectral coefficients (MFSC), to the decorrelated mel-frequency cepstral coefficients (MFCC). MFSC features make it easier for DNNs to discover linear relations as well as higher order causes of the input data, leading to better overall system performance. Hence we do not consider MFCC features in our experiments and use the filter bank features instead.

The second challenge arises due to the larger number of parameters that a model needs when handling the audio information. The number of parameters in the Video-Text baseline is 16.95 million as compared to 32.08 million when we add audio. This is because of the high number of input timesteps in the audio modality encoder, which makes learning trickier and more time-consuming.

To demonstrate these challenges, as an experiment, we group the audio features across input timesteps into bins with an average of 30 consecutive timesteps and train our MAST model. This makes the number of audio timesteps comparable to the number of video and text timesteps. While we observe an improvement in computational efficiency, it achieves a lower performance than the baseline Video-Text model as described in Table \ref{tab:results} (MAST-Binned). We also train Audio only and Audio-Text models which fail to beat the Text only baseline. We observe that the generated summaries of the Audio only model are similar and repetitive, indicating that the model failed to learn useful information relevant to the task of text summarization.

\section{Results and Discussion}


\begin{table}[!htbp]
    \centering
    \footnotesize
    \begin{tabular}{|c|c|c|c|c|}
        \hline
        \textbf{Model} & \multicolumn{3}{|c|}{\textbf{ROUGE}} & \textbf{Content}\\
        \cline{2-4}
        \textbf{Name} & \textbf{1} & \textbf{2} & \textbf{L} & \textbf{F1} \\
        \hline
        Text Only & 46.01 & 25.16 & 39.98 & 33.45 \\
        BertSumAbs & 29.68 & 11.74 & 22.58 & 31.53 \\
        Video Only & 39.23 & 19.82 & 34.17 & 27.06 \\
        Audio Only & 29.16 & 12.36 & 28.86 & 26.65 \\
        Audio-Text & 34.56 & 15.22 & 31.63 & 28.36 \\
        Video-Text & 48.40 & 27.97 & 42.23 & 32.89 \\
        TrimodalH2 & 47.85 & 28.46 & 42.17 & \textbf{35.65} \\
        MAST-Binned & 46.22 & 25.94 & 40.34 & 33.56 \\
        MAST & \textbf{48.85} & \textbf{29.51} & \textbf{43.23} & 35.40 \\
        \hline
    \end{tabular}
    \caption{Results for different configurations. MAST outperforms all baseline models in terms of ROUGE scores, and obtains a higher Content-F1 score than all baselines while obtaining a score close to the TrimodalH2 model.}
    \label{tab:results}
    \vspace{-3.5mm}
 \end{table}

\subsection{Preliminaries}
Our results are given in Table \ref{tab:results}. To demonstrate the contribution of various modalities towards the output summary, we experiment with the three modalities taken individually as well as in combination. Text only, Video only and the Audio only are attention-based S2S models \cite{bahdanau2014neural} with their respective modality features taken as encoder inputs. To situate the efficacy of the encoder-decoder architecture for our task, we use the BertSumAbs \cite{liu2019text} as a BERT based baseline for abstractive text summarization. Audio-Text and the Video-Text are S2S models with hierarchical attention layer. The Video-Text model as presented by \citet{palaskar2019multimodal} has been compared on the 300h version instead of the 2000h version of the dataset because the audio modality is only available in the former. TrimodalH2 model, adds the audio modality in the second-level of hierarchical attention. MAST-Binned model groups the features of the audio modality for computational efficiency. These models show alternative methods for utilizing audio modality information.

We evaluate our models with the ROUGE metric \cite{lin2004rouge} and the Content F1 metric \cite{palaskar2019multimodal}. The Content F1 metric is the F1 score of the content words in the summaries based on a monolingual alignment. It is calculated using the METEOR toolkit \cite{denkowski2011meteor} by setting zero weight to function words ($\delta$), equal weights to Precision and Recall ($\alpha$), and no cross-over penalty ($\gamma$) for generated words. Additionally, a set of catchphrases like the words - in, this, free, video, learn, how, tips, expert - which appear in most summaries and act like function words instead of content words are removed from the reference and hypothesis summaries as a post-processing step. It ignores the fluency of the output, but gives an estimate of the amount of useful content words the model is able to capture in the output.

\subsection{Discussion}
As observed from the scores for the Text Only model, the text modality contains the most amount of information relevant to the final summary, followed by the video and the audio modalities. The scores obtained by combining the audio-text and video-text modalities also indicate the same. The transformer-based model, BertSumAbs, fails to perform well because of the smaller amount of text data available to fine-tune the model.

We also observe that combining the text and audio modalities leads to a lower ROUGE score than the Text Only model, which indicates that the plain hierarchical attention model fails to learn well over the audio modality by itself. This observation is in line with the result obtained by the TrimodalH2 model, where we simply extend the hierarchical attention approach to three modalities.

\subsubsection{Usefulness of audio modality}
 
The MAST and the TrimodalH2 models achieve a higher Content F1 score than the Video-Text baseline, indicating that the model learns to extract more useful content by utilizing information from the audio modality corresponding to the characteristics of speech, in line with our initial hypothesis as illustrated in Table \ref{tab:contents}

However, the TrimodalH2 model, which simply adds the audio modality in the second level of hierarchical attention, fails to outperform the Video-Text baseline in terms of ROUGE scores. Our architecture lets the MAST model choose between paying attention to a different combination of modalities with the text modality. This forces the model to pay more attention to the text modality, thereby overcoming the shortcoming of the TrimodalH2 model and achieving better ROUGE scores, while maintaining a similar Content F1 score when compared to TrimodalH2.

\begin{figure}
    \centering
    \includegraphics[scale=0.33]{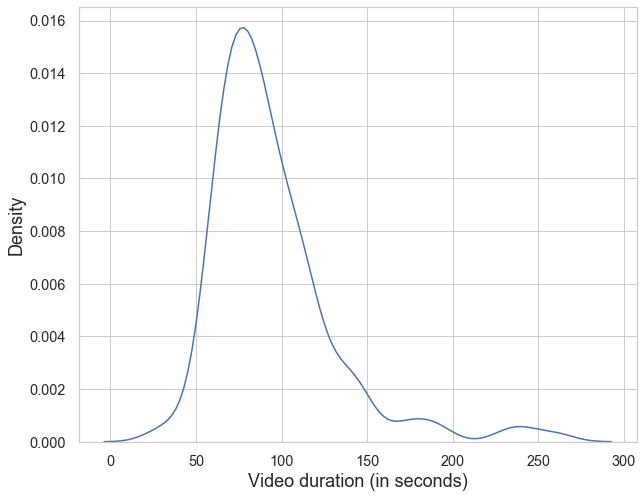}
    \caption{Distribution of the duration of videos (in seconds) in the test set.}
    \label{fig:duration_dist}
    \vspace{-3.5mm}
\end{figure}

\subsubsection{Attention distribution across modalities}
\begin{figure*}[!ht]
  \begin{subfigure}{0.333\textwidth}
    \centering
    \includegraphics[width=0.99\linewidth, trim=5 0 30 20, clip]{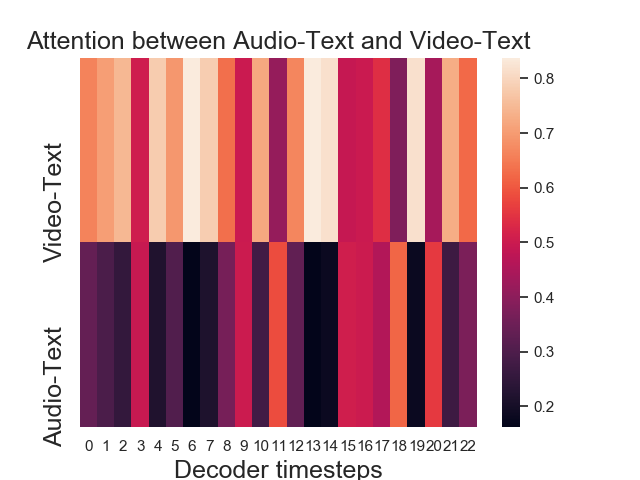}
    \caption{}
    \label{fig:vis1}
  \end{subfigure}%
  \begin{subfigure}{0.333\textwidth}
    \centering
    \includegraphics[width=\linewidth, trim=5 0 30 20, clip]{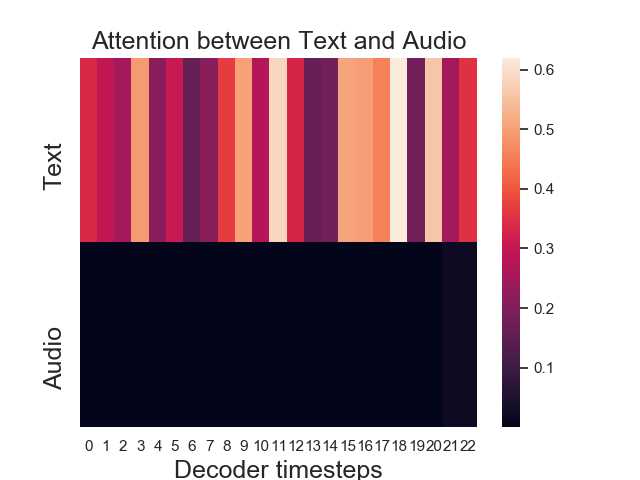}
    \caption{}
    \label{fig:vis2}
  \end{subfigure}
  \begin{subfigure}{0.333\textwidth}\quad
    \centering
    \includegraphics[width=0.97\linewidth, trim=5 0 45 20, clip]{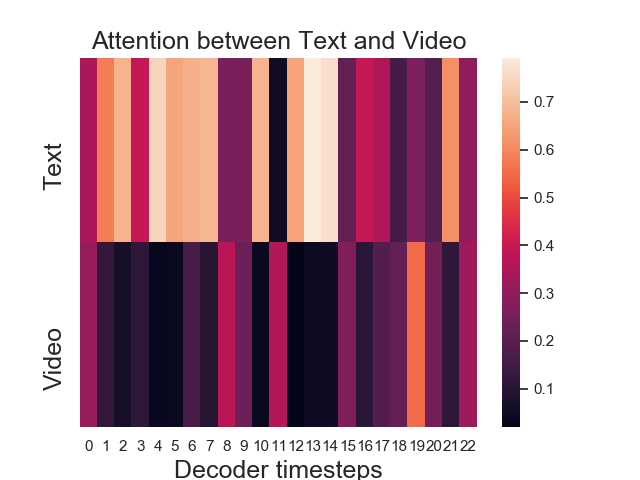}
    \caption{}
    \label{fig:vis3}
  \end{subfigure}
  \medskip

  \begin{subfigure}{0.33\textwidth}
    \centering
    \includegraphics[width=0.99\linewidth, trim=5 0 30 20, clip]{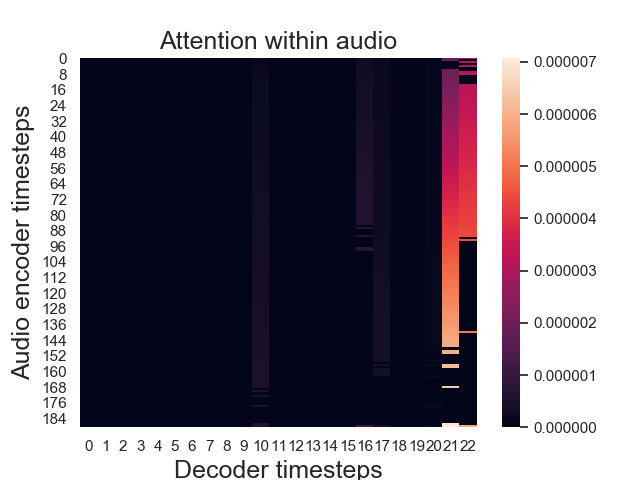}
    \caption{}
    \label{fig:vis4}
  \end{subfigure}
  \begin{subfigure}{0.333\textwidth}
    \centering
    \includegraphics[width=\linewidth, trim=5 0 30 20, clip]{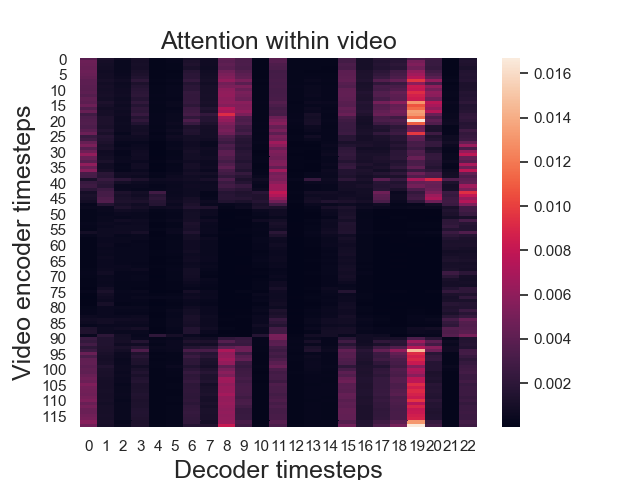}
    \caption{}
    \label{fig:vis5}
  \end{subfigure}
  \begin{subfigure}{0.333\textwidth}
    \centering
    \includegraphics[width=0.97\linewidth, trim=5 0 45 20, clip]{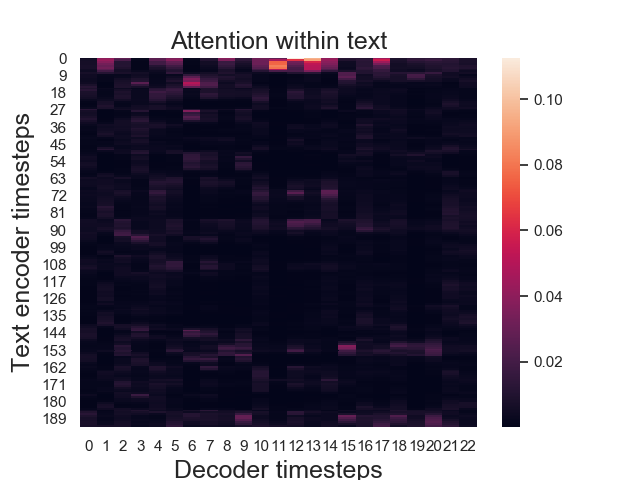}
    \caption{}
    \label{fig:vis6}
  \end{subfigure}
  \medskip
  \caption{Visualization of attention weights in the Trimodal Hierarchical Attention layer for a sample video in the test set. Figures \ref{fig:vis1} to \ref{fig:vis3} show the varying attention distribution on different combinations of modalities across the decoder timesteps. Figures \ref{fig:vis4} to \ref{fig:vis6} show the attention distribution on the encoder timesteps for each modality across the decoder timesteps. This shows the usefulness of each modality for the generation of the summary.
  }
  \label{fig:attention_vis}
  \vspace{-2.5mm}
\end{figure*}

To understand the importance of individual modalities and their combinations, we plot their attention distribution at different levels of attention hierarchy across the decoder timesteps. Figure \ref{fig:vis1} corresponds to attention weights as calculated in equation \ref{eq14} while figures \ref{fig:vis2} and \ref{fig:vis3} correspond to the product of attention weights between equations \ref{eq11}, \ref{eq8} and corresponding weight in equation \ref{eq14} for each decoder timestep. The final attention within each individual modality at each decoder timestep is calculated by multiplying the corresponding cumulative attention weights obtained at level 2 of attention hierarchy with the attention weights obtained in equation \ref{eq2} (figures \ref{fig:vis4} to \ref{fig:vis6}). The attention weights assigned to the audio modality have been added across input timesteps (group size of 30) in order to obtain a more interpretable visualization.

Through these visualizations, we observe that the text modality dominates the generation of the output summary while giving lesser attention to the audio and video modalities (the latter being more important). These findings support the extra importance being given to the text modality in the MAST model during its interaction with the other modalities. Figures \ref{fig:vis2} and \ref{fig:vis4} highlight the modest gains through the audio modality and the challenge in its appropriate usage.

\subsubsection{Performance across video durations}

\begin{figure}
    \centering
    \includegraphics[scale=0.34]{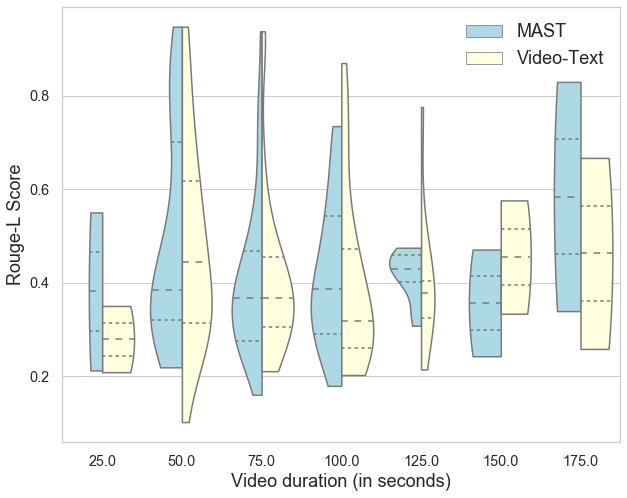}
    \caption{Distribution of Rouge-L scores of summaries produced for different video durations (in seconds) for MAST and Video-Text baseline. The videos are binned into groups of 25 seconds by duration and the distribution of Rouge-L scores within each group is shown using density plots. The dotted lines inside each group show the quartile distribution.}
    \label{fig:rouge_dist}
    \vspace{-3.5mm}
\end{figure}

We also look at how our model performs for different video durations in our test set. Figure \ref{fig:rouge_dist} shows the variation in the Rouge-L scores across different videos for MAST and the Video-Text baseline. The figure shows videos binned into seven groups of 25 seconds by duration. We can observe from the quartile distribution that MAST outperforms the baseline in five out of the seven groups, gives similar performance for videos with a duration between 75-100 seconds, and underperforms for videos with a duration between 150-175 seconds. However, overall, by looking at the distribution of the duration of videos in our test set (Figure \ref{fig:duration_dist}), we can observe that MAST outperforms the baseline for a vast majority of videos across durations.


\section{Related Work}

\subsection{Abstractive text summarization}
Abstractive summarization of documents was traditionally achieved by paraphrasing and fusing multiple sentences along with their grammatical rewriting \cite{woodsend2012multiple}. This was later improved by taking inspiration from human comprehension capabilities when \citet*{fang2014summariser} implemented the model of human comprehension and summarization proposed by \citet*{kintsch1978toward}. They did this by identifying these concepts in text through the application of co-reference resolution, named entity recognition and semantic similarity detection, implemented as a two-step competition.

The real stimulus to the field of abstractive summarization was provided by the application of neural encoder-decoder architectures. \citet{rush2015neural} were among the first to achieve state-of-the-art results on Gigaword \cite{graff2003english} and the DUC-2004 \cite{over2007duc} datasets and established the importance of end-to-end deep learning models for abstractive summarization. Their work was later improved upon by \citet{see2017get} where they used copying from the source text to remove the problem of incorrect generation of facts in the summary, as well as a coverage mechanism to curb the problem of repetition of words in the generated summary.

\subsection{Pretrained language models}
Another breakthrough for the field of natural language processing came with the use of pre-trained language models for carrying out various language downstream tasks. Pre-trained language models like BERT \cite{devlin2018bert} introduced masked language modelling, which allowed models to learn interactions between left and right context words. These models have significantly changed the way word embeddings are generated by training contextual embeddings rather than static embeddings. \citet{liu2019text} presented how BERT could be used for text summarization and proposed a new fine-tuning schedule for abstractive summarization which adopted different optimizers for the encoder and the decoder to alleviate the mismatch between the two. BERT models typically require large amounts of annotated data to produce state-of-the-art results. Recent works, like GAN-BERT by \citet{croce2020gan} focus on solving this problem.

\subsection{Advancements in speech recognition and computer vision}
Parallel advancements in the field of speech recognition and computer vision have been able to give us successful methods to extract useful features of speech and images. \citet{peddinti2015jhu} built a robust acoustic model for speech recognition using a time-delay neural network. They were able to achieve state-of-the-art results in the IARPA ASpIRE Challenge. Similarly, with the advancements of convolutional neural networks, the field of computer vision has progressed significantly. \citet{he2016deep} demonstrated the strength of deep residual networks which learned residual functions with reference to the layers and were able to achieve state-of-the art results on the ImageNet dataset. \citet{hara2018can} showed that simple 3D Convolutional Neural Network (CNN) architectures outperform complex 2D architectures and trained a ResNeXt-101 3D CNN to recognize 400 different human actions on the Kinetics dataset \cite{kay2017kinetics}.

\subsection{Summarization beyond text}
The advancements in these fields have in turn also facilitated text summarization. \citet{rott2016speech} used only the input audio to generate textual summaries while \citet{sah2017semantic} were among the first to show the possibility of summarizing long videos and then annotating the summarized video to obtain a textual summary. These models, however, were not able to capture the information of other modalities to obtain the output textual summary and hence their limitations led to the increasing use of multimodal data. A major hindrance in the field of multimodal text summarization was the lack of datasets. \citet{li2017multi} created an asynchronous benchmark dataset with human-annotated summaries for 500 videos. \citet{sanabria2018how2} then released a large-scale dataset for instructional videos. \citet{jn2020multimodal} and \citet{zhu2018msmo} presented multimodal text summarization models using textual and visual modalities as input and multimodal outputs of summarized text and video. \citet{palaskar2019multimodal} used How2 dataset to present an abstractive summary of open-domain videos. These models, however, are not completely multimodal since they do not utilise the audio information. A major focus of our work is to highlight the importance of using audio data as input and incorporate it in a truly multimodal manner.

\section{Conclusion}
In this work\footnote{\url{https://github.com/amankhullar/mast}}, we presented MAST, a state of the art sequence to sequence based model that uses information from all three modalities -- audio, text and video -- to generate abstractive multimodal text summaries. It uses a Trimodal Hierarchical Attention layer to utilize information from all modalities. We explored the role played by adding the audio modality and compared MAST with several baseline models, demonstrating the effectiveness of our approach.

In the future, we would like to extend this work by looking at alternate audio modality representations including using neural networks for audio feature extraction, and also explore the use of transformers for an end to end attention based learning. We also aim to explore the application of MAST to other multimodal tasks like translation.



\bibliography{refs}
\bibliographystyle{acl_natbib}

\end{document}